# Unsupervised Multidomain Approaches to Named Entity Recognition with Small Datasets

**Israel Fianyi[a]. James Montgomery[b]. Soonja Yeom[b]**

[a] School of Information Communication Technologies, University of Tasmania, Launceston, Tasmania, Australia
[b]School of Information Communication Technologies, University of Tasmania, Hobart, Tasmania Australia
***Corresponding author. Email: Israel.fianyi@utas.edu.au**

**Abstract**

This paper explores the challenges and the methodologies associated with learning quality representations in scenarios with unlabelled small or limited datasets for downstream information extraction task (Multidomain Named Entity Recognition (NER). The study adopts a Transfer Learning on small datasets. Traditional NER systems often rely on large, labelled data, which is impractical for many domains. This study, therefore, applies an unsupervised pre-training approach to precondition and identify entities without annotated datasets, then applies transfer learning models to different simulated limited datasets for a named entity recognition task. Entity Recognition (NER) is essential in natural language processing (NLP), it identifies and classifies related entities within the text. This study addresses the complexities of domain variability, data sparsity, and overfitting and investigates innovative approaches such as data augmentation, few-shot learning, and domain adversarial training. Integrating these techniques promises to enhance the performance and generalizability of NER systems across diverse and resource-constrained domains, paving the way for more efficient and adaptable NLP applications.

Keywords: Unsupervised learning, Multimodal, Named entity Recognition, Transfer Learning, Information Extraction

# Introduction

Named Entity Recognition (NER) is a fundamental task in Natural Language Processing (NLP) that involves identifying and classifying entities such as names, organisations, locations and other proper nouns within a text (Qu, Gu, et al., 2023). Traditionally, NER systems rely heavily on supervised learning methods, which require large, annotated datasets to achieve high accuracy and performance. However, obtaining such extensive labelled data is often impractical, especially when dealing with multiple domains or specialised fields where annotated resources are scarce (Mengliev et al., 2024; Qu, Zeng, et al., 2023).

The landscape of NER is evolving with the advent of unsupervised learning and transfer learning techniques, which offer promising alternatives to traditional methods. Unsupervised NER aims to identify entities without the need for labelled training data, thus making it more adaptable to various domains and reducing the dependency on extensive human annotation. On the other hand, transfer learning leverages pretrained models trained on large general datasets, enabling them to be fine-tuned for specific tasks or domains with minimal additional data. However, there are certain domains and specialised areas that would not have large datasets

Combining these approaches-unsupervised learning and transfer learning, holds significant potential for advancing NER, particularly in multidomain settings where data variability and scarcity pose significant challenges. This research focuses on the intersection of these methodologies, aiming to develop robust NER systems that can operate effectively across diverse domains using small datasets. Key challenges include handling domain variability, mitigating overfitting, and ensuring efficient knowledge transfer.

In this study, we explore various innovative techniques, including data augmentation, few-shot learning, and domain-adversarial training, to enhance the performance and generalizability of NER systems. By addressing these challenges, our research seeks to contribute to the development of more adaptable, efficient, and accurate NER applications, ultimately broadening the scope and utility of NLP technologies in real-world scenarios where labelled data is limited.

This study empirically explores the detection of entity mentions from small unlabelled text corpus into clusters for an unsupervised Named Entity Linking (NEL). Every deep learning mechanism for NEL requires exploration of unknown structures in the input distribution in source domain to discover quality representation for a target domain's data (Cao et al., 2019). Unsupervised transfer learning and unsupervised pretraining techniques are some of the recent approaches to exploiting input distributions from unlabelled text data. The conventional approach for unsupervised pretraining algorithms and then applying the learning onto a new dataset (labelled or semi-labelled) has been underpinned by a large dataset (for most transfer learning task), which has been successful (Bengio, 2012; Cao et al., 2019; You, Liu, Wang, & Long, 2021). However, little work has been done to exploit a fully unlabelled small dataset in a transfer learning approach to learn representations for NEL, an NLP task. Unsupervised transfer learning with small datasets is mostly overfitting (Arnold et al., 2007; Raffel et al., 2020). The investigation in this section proposes a new Unsupervised Transfer Learning Approach (UTL) with a small dataset in learning representations for a downstream unsupervised Named Entity Linking.

The study investigates how combining unsupervised pretraining techniques, and a transfer learning approach facilitates good word embeddings for unsupervised named entity Linking with a small dataset. Attaining this objective helps address the proposed research question: *To what extent will joint application of unsupervised pretrained algorithms and unsupervised transfer learning on a different dataset in the original dataset facilitate quality Open Named Entity-Linking from the text?* Figure 1 illustrates the proposed transfer learning techniques.

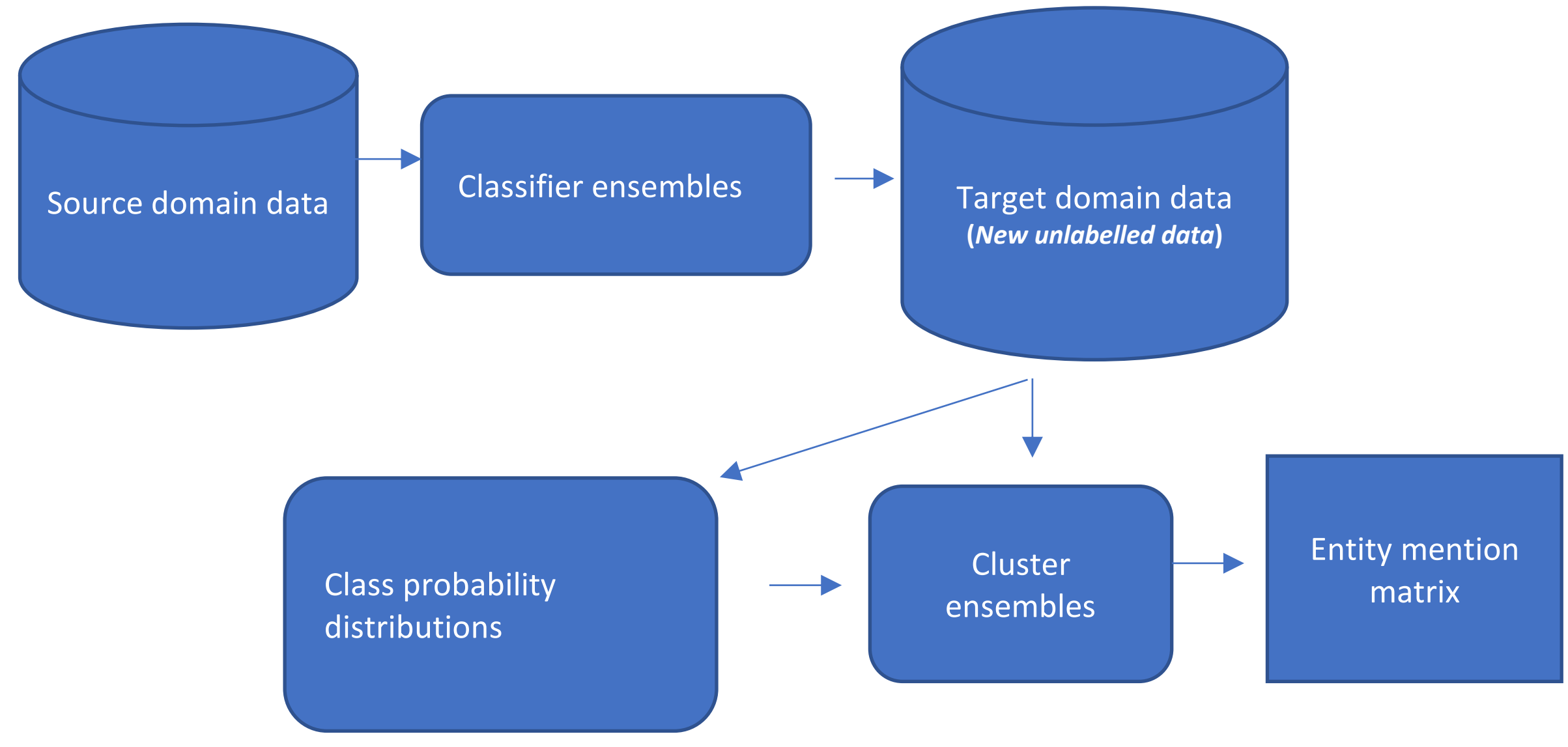

Figure 1. Overview of the proposed transfer learning technique

The experimental design for the unsupervised transfer learning for named entity linking experiment is divided into three phases: In the first stage of unsupervised transfer learning, all the layers are initiated with an unsupervised pretraining indicator using greedy layer-wise training. During the second phase of the experiment, a universal training standard is minimised while exploring the several use cases of the training design. The third phase compares the quality of the proposed model for names entity links against existing state-of-the-art methods.

# Methodologies, Model Training and Architecture

The structure of all the methodological approaches used for this study are organised as follows:

## 2.1.1 Greedy Layer-Wise Learning of Representations

While greedy layer-wise learning was introduced in 2007 by Bengio, Lamblin, Popovici, and Larochelle (2007) and has since been the foundation for the most recent advanced representation learning algorithm with large datasets, the proposed method for this study departs from the traditional approach. It adopts the following convention to learn from a small dataset.

1. Let $h_r(x) = x$ represent the low-level representation of the dataset while the approach observes the input x.

2. Furthermore, let $\ell = 1$; where an unsupervised training models $h_{\ell-1}$ and embedded at $\ell - 1$, are observed, as it generates a representation

3. $h_{x(\ell)} = R_{\ell}(h_{\ell-1}(x))$ The out output for the next level.

Several alternatives in the literature have been explored, and it is worth noting that supervised learning representation is more dominant than unsupervised representation learning for small datasets. As such, for this study, the unsupervised learning model based on the parameters is initialised at the first stages.

4. The investigation then finetunes the unsupervised learning model in a training loss with the given training set.

As mentioned in Smalheiser, Cohen, and Bonifield (2019), training works better in a low-dimension space, that is, from the input to the classifier, regardless of the size of the dataset, mostly between 50 to 500 dimensions. The significance of this is the difficulty associated with interpreting higher dimensions meaningfully in NLP.

### 2.1.2 Transfer Learning and Domain Adaptation in Transductive Settings

Given that the training is solely on unlabelled datasets for the NEL task, most supervised fine-tuning mechanisms used in transfer learning and domain adaptation are not possible in this case. Consequently, for the proposed transfer learning method with unlabelled data, the input distribution is different between the training data and the test set. Most of the training examples are different from those in the test set examples in terms of the clusters or classes. While this is an extreme transfer learning and domain adaptation setup, the study generalises the approach in context so that the training set for the input distribution has nothing to do with the test set.

Subsequently, the representation learned captures the non-specified factors that enables variations in all the respective classes. Consequently, the CNN classifier trained on the test set captured those factors imperative to the discriminant in the test set classes.

A preliminary investigation revealed that due to the small size of the unlabelled examples, to validate the classifier, the investigation could not immediately achieve good performance on the test set with high dimensions. As a result, the proposed method selected the relevant features using a transductive approach (Boudiaf et al., 2021)[1]. The top level of the unsupervised learning hierarchy layer is purely trained on the test examples. This is typically achieved by applying Principal Component Analysis (PCA) at the last level of the hidden layer, where the study trains only on test sets to select a few eigenvectors.

[1] The proposed unsupervised learning algorithm builds a generic model from the observed training and test sets

### 2.1.3 Principal Component Analysis, Independent Component Analysis

Principal Component Analysis and Independent Component Analysis are linear models that successfully support several levels of deep hidden layers (Fan & Wang, 2014; Wetzel, 2017). The PCA conserves the global linear directions at the highest variance, segregating them into rectangular components. The projection of the main eigenvectors and the input covariance matrix determine the representation learned. Interestingly, several of the challenges of working with small datasets make using PCA the initial and final level, the appropriate algorithms for dimension reduction in this investigation. For the first layer of the PCA, the study keeps a minimal direction against the traditional approach that uses many directions. This step aims to smoothen the input distribution and eliminate some of the variations in the different global directions. Notably, PCA can perform normalisation across samples by subtracting the mean sample and dividing it by the stand deviation in the chosen direction. The investigation also employs the *contrast normalisation* (Ashraf et al., 2013) variant as an intermediate step in the deep Convolutional Neural Network used for the unsupervised Named Entity Linking model of this study.

### 2.1.4 Convolutional Denoising Autoencoders

Recent advances in denoising autoencoder focus on dataset challenges where the input vectors are extremely large and sparse (Duong et al., 2022; Huang & Sun, 2016; Mesnil et al., 2012; Zhu et al., 2022). The traditional autoencoders for transfer learning map the input $x \in \mathbb{R}^{dx}$ into a latent representation denoted as *h* for features learnt, with non-linearity sigmoid *s:*

$$h = f(x) = s\,(Q_x + b), \quad [1]$$

*where Q is the tied weight*.

The representation learned is mapped back into the input space. The denoising autoencoder is trained to denoise and not merely train to reconstruct the inputs artificially. The training in these existing approaches requires large datasets so that the encoder is presented with unadulterated sample ***s***, with an input given stochastically ***ś*** as a corrupted version. The challenge here is that when a small dataset sample is introduced, the model performs poorly, implying that this model cannot be used in scenarios where there is not much data. The proposed method focuses on training convolutional denoising autoencoders (CoDAE) on small, sparse vectors, where the encoder takes zeros in the input vector without any further computation to learn the entity embedding. The proposed method uses the decoder to reconstruct the entity text segmentation and compute the reconstruction error with all the inputs, including the zeros for the entity mention clusters. Building upon the work of Wan et al. (2021), the experiment uses 10-fold cross-validation techniques while the study further reconstructs a small stochastically designated and reconstruction error, including the zeros.

The investigation adopts a constructive autoencoder to enhance the robustness of the learned representation *f(x)* from the input training set *x*, the (Dauphin et al., 2012; Taniguchi, Nakashima, Liu, & Nagasaka, 2016) to correct the sensitivity associated with the small input training set using *Jacobian* $J_f(x)$ non-linear map of the Frobenius norm as proposed in Shi, Lei, Ma, and Niu (2019).

For this experiment, given the input $x \in \mathbb{R}^{dx}$, mapped into the encoder function *f,* the latent representation $h \in \mathbb{R}^{dh}$, where the sensitivity control correction is the total sum of squares of the entire partially generated derivatives of the features extracted with regards to the small set input dimension is defined as:

$$||J_f(x)||_f^2 = \sum_{rj}\left(\frac{\partial h_j(x)}{\partial x_r}\right)^2 \qquad [2]$$

Where $||J_f(x)||_f^2$ facilitates constructive mapping in the neighbourhood of the training data in a respective feature space. The small initial derivative denotes the *robustness* of the representation learned from the small variational input.

In monitoring the performance of the proposed algorithm during training, the convolutional denoising reconstruction error is an excellent measure of the progress of the model because it corresponds with the training criterion and can help in early stopping. Notably, the best standalone denoising mechanism does not always work well when providing a quality representation for a classifier. The approach is consistent in all transfer learning settings, usually where training distribution is different from the test and the validation distribution. As a result, the convolutional denoising reconstruction error is the best-chosen solution for evaluating the validation cluster error at a diverse training time frame. The experiment did not stop training but recorded the representation learned at different stages of the training trajectory. The investigation did not retrain a separate model from scratch for each duration tested.

### 2.1.5 Hyperparameters, Architecture and Parameters

The most vital hyperparameter for all the algorithms used in this experiment is the learning rate. When the learning rate is low, the approach has slow convergence, and when the learning rate is large, it produces poor performance due to the increase in the training criterion. The optimal learning rate for the derived small dataset is small enough when changing, which helps optimise the learning rate. Like most numerical hyperparameters that propose varied learning rates, this investigation explores the log domain with the dynamic range.

The Convolutional Denoising Autoencoders (CoDAE) is devised to provide a small set of representations for a classification activity. In this investigation, the approach is designed with two convolutional denoising autoencoders of the architecture and parameters, as shown in

Table *1* and

Table *2*. The encoder comprises 1-dimensional convolution and 1-dimensional max-pooling layers stacked on top of each other. The proposed method uses a kernel size of 11 for the initial 1-dimensional convolutional layer as inspired by various studies (Chadha et al., 2021; Niepert et al., 2016; Shi et al., 2021), that used this approach. Notably, the kernel size decreases as the network expands to the final 1-dimensional convolutional layer with three kernels denoted as **A** and 6 for the convolutional denoising autoencoders denoted as **B**. The output shape is the same as the shape of the input; subsequently, the investigation reduces the output size by half resulting in the max-pooling. The Convolutional layers also have a representation map of 10 as it progressively increases to 50 in **A** and 40 in **B** of the 1-dimensional convolutional denoising autoencoder. The encoder representation at the encoding layer determines the shape of the output and is determined by the 1-dimensional max-pooling operators, including the parameters of the 1-dimensional convolution.

Table 1. The Convolutional Denoising Autoencoder architecture and parameters **[A]**: The layer 1 to 10 denotes the encoder. The $11^{th}$ layer is the encoding layer. The decoder is symmetrical to the encoder in terms of the structure of the hidden layer.

| | Kernel Size | Layer | Stride | Shape of the output |
|---|---|---|---|---|
| 1 | input | | | 100 x 3 |
| 2 | Conv | 10 | 1 | 100 x 10 |
| 3 | Max-pooling | 2 | 2 | 50 x 10 |
| 4 | Conv | 8 | 1 | 50 x 20 |
| 5 | Max-pooling | 2 | 2 | 25 x 20 |
| 6 | Conv | 5 | 1 | 25 x 30 |
| 7 | Max-pooling | 2 | 2 | 12 x 30 |
| 8 | Conv | 3 | 1 | 3 x d |
| 9 | Conv | 3 | 1 | 3 x 50 |
| 10 | up sample | 2 | 2 | 6 x 40 |
| 11 | Conv | 10 | 1 | 12 x 30 |

Table 2. The Convolutional Denoising Autoencoder architecture and parameters **[B]**: The layer 1 to 9 denotes the encoder. The 10th layer is the encoding layer. The decoder is symmetrical to the encoder in terms of the structure of the hidden layer.

| | Kernel Size | Layer | Stride | Shape of the output |
|---|---|---|---|---|
| 1 | input | | | 100 x 3 |
| 2 | Conv | 10 | 1 | 100 x 10 |
| 3 | Max-pooling | 2 | 2 | 50 x 10 |
| 4 | Conv | 8 | 1 | 50 x 20 |
| 5 | Max-pooling | 2 | 2 | 25 x 20 |
| 6 | Conv | 6 | 1 | 25 x 30 |
| 7 | Max-pooling | 2 | 2 | 12 x 30 |
| 8 | Conv | 3 | 1 | 3 x d |
| 9 | Conv | 3 | 1 | 3 x 40 |
| 10 | up sample | 2 | 2 | 6 x 40 |

### 2.1.6 Extracting Named Entity Mentions

The methodology presented in the preceding section focuses on various approaches in learning quality representation from small datasets for a downstream task of unsupervised named entity linking. The methodology encompasses the adaption greedy layer-wise pretraining techniques to pretrain the small datasets. Then the study introduces the transfer and domain adaptation techniques to use the embedding model to learn representations for named entity linking task later. Furthermore, the study proposes the convolutional denoising autoencoders to support the transfer learning approach, while at the same time using it as the classifier for the named entity linking task. In the next section, the evaluate the proposed methodology on unsupervised named entity linking from small unlabelled dataset.

The experiment formalises a multiclass clustering problem to detect entity mentions while mapping words and knowledge base resources in similar representation space. The encoder of the autoencoder compresses the input data by spreading the data from one hidden layer to the other. The decoder of the autoencoder takes on the encoded representation and then reconstructs the small input data. Subsequently, the study uses stochastic noise to reconstruct the data from the distorted data for a more effective segregation of closely linked representations learned. The approach uses the 1-dimensional convolution and 1-dimension max-pooling to build the denoising encoder to facilitate the salient representation learning for more robust extraction of related entity mentions. Figure 2 Shows convolutional denoising autoencoder (CoDAE) architecture for the proposed transfer learning approach.

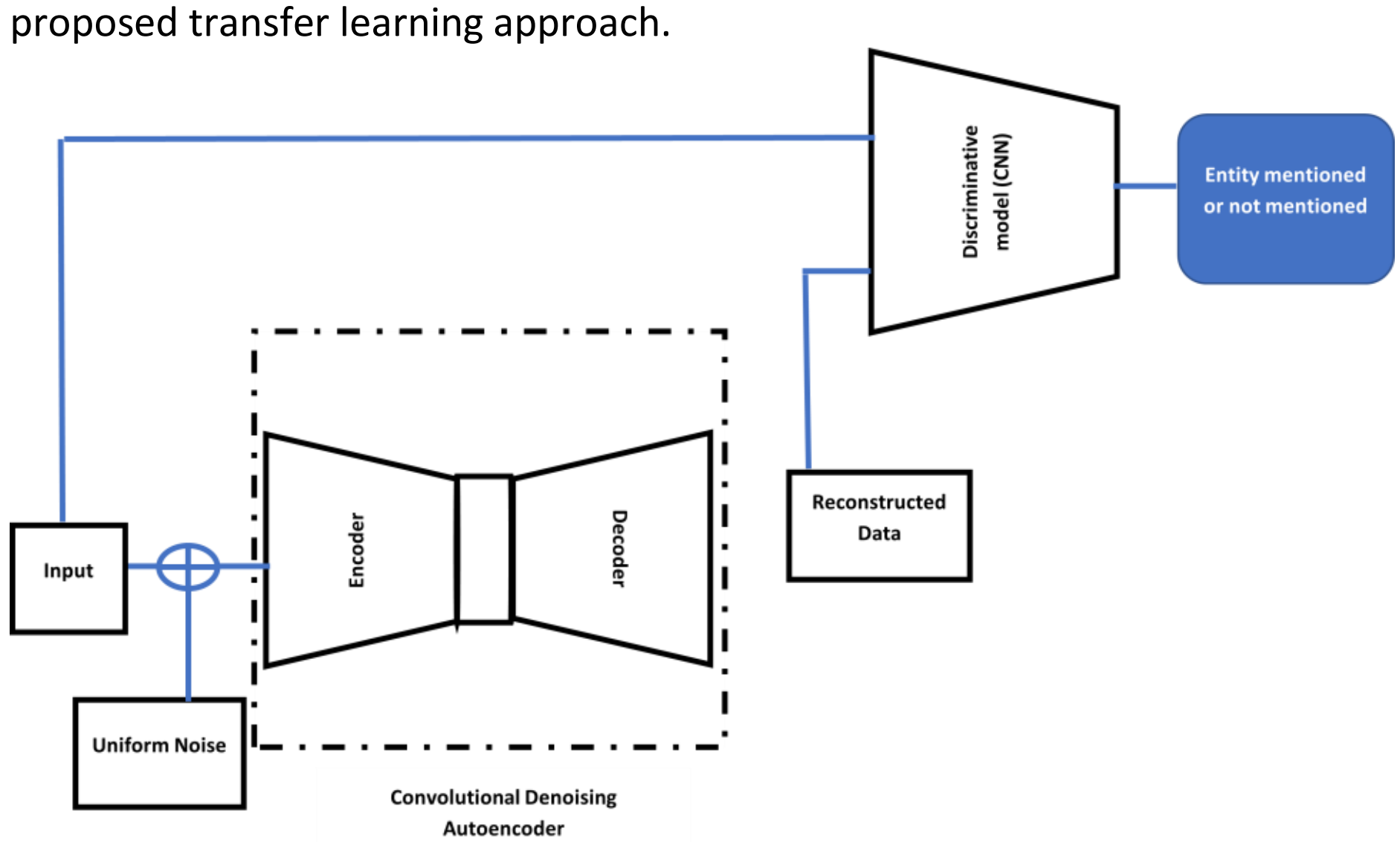

Figure 2. The Proposed Convolutional Denoising Autoencoder Architecture

## 2.2 Experiments: Evaluation of the Models

### 2.2.1 Datasets

The *source domain* for this evaluation is Reuters-21578, the choice of this dataset is because of its composition and it consist of the ground truth for the named entity linking task.: Firstly, the study pretrains the model on Reuters-21578 as the source domain data.[2] It contains 10,788 documents of the Reuters newswire service. The dataset has no labels and is partitioned with 7769 training sets and 3019 test sets. The sentences in all the 21 files in Reuters-21578 are not annotated with predetermined pair of entities and their related classes. There are 14 ordered relations (single direction) and undirected *other* clusters culminating into 15 clusters. A paired entity is classified as correct if the order of the related entities in the cluster is accurate.

[2] http://kdd.ics.uci.edu/databases/reuters21578/reuters21578.html

For the *target domain*: Automatic Content Extraction (ACE)2005: The transfer learning evaluation in this experiment is performed on ACE2005[3] dataset. ACE2005 consists of a multilingual corpus, and the genre incorporates broadcast news, newswire, blogs, and forums. However, for this experiment, the study uses only the English language corpus, Table 3 shows the distribution of the ACE2005 datasets. The ACE was developed to facilitate the automatic extraction of relevant content from human language. In November 2005, it was used to evaluate the performance of several NLP and information extraction problems. Therefore, the proposed method uses the first pass subject dataset without the labels to create a scenario of evaluating the proposed model on a small dataset. The proposed method used only 20% (60767 words) of the tokenise corpus without the labels. The use of 20% of the dataset was inspired by Tekumalla and Banda (2021), who used weakly supervised techniques to generate a percentage of training data from the large social media text corpus. The study uses this dataset because the words and knowledge base resources must be in the same representation space to transfer a word embedding model, and the ACE2005 corpus has both. The structure of the ACE2005 dataset sufficiently models the requirement of the proposed method since the named entity can directly be associated with the dataset. To avoid bias in the unlabelled datasets, the study uses the unsupervised neural network to detect the bias trend. Consequently, the input data should be different from the output result. Another bias prevention factors the study adopted was to use datasets that replicates real-world scenarios and not to reuse the datasets.

Table 3. Distribution of the English Language Corpus of the ACE2005 dataset

| English Language Corpus | | | |
|---|---|---|---|
| Words | | | |
| First pass data subject | Dual first pass | Adjudication | TIMEX2 normalization |
| 303833 | 297185 | 216545 | 259889 |

[3] https://catalog.ldc.upenn.edu/LDC2006T06

- Evaluation of the proposed method is carried out on two benchmark domain-theory datasets from the University of California machine learning repository[4].
    1. Text preprocessing (TERRY)[5] (Hemavathi et al., 2019)
    2. Health News in Twitter Data Set[6] (Dai et al., 2017)

### 2.2.2 Transfer Learning with Clustering Algorithm

Given the unlabelled input model $K^s$ drawn from the pretrained source input space ***S*** and the target data ***T*** where S ≠ T In the proposed transfer cluster model, no annotated class on the source and target text data is used. The aim is to ascertain how the joint application of unsupervised pretraining and transfer learning techniques on a small dataset can facilitate learning of sufficiently good representation on a target sample $f^{\theta}: q^t \rightarrow y^v$, resulting in a conditional probability that each of the output of the training samples belongs to each $K^s$ cluster $f^{\delta}: y^v \rightarrow p$, so that the target sample is in the same cluster with the same representations.

The investigation constructs the initial cluster solution by employing the unsupervised pretraining model on the source data to transfer prior knowledge. The transfer learning model is built on the understanding that there is shared common knowledge between the target data and the source domain or the ground-truth data. Consequently, yielding an initial representation of the target data $T^d$ by feeding them into:

$$f_{\theta}: \acute{y}^t = \acute{f}_{\theta}\left(T^d\right) \qquad [3]$$

Where initial clusters of $\acute{y}^t$ is computed using the K-means algorithm, while the proposed method separates $\acute{y}^t$ into $K^s$, with the notion that the source data contains prior knowledge applicable to the target data. Conspicuously, this notion did not always work as expected due to the discrepancies in the data, including continuous distributional shift, during the initial investigation. Therefore, the study uses a denoising feature learning strategy to mitigate this challenge and enhance domain adaptation.

### 2.2.3 Denoising Features for Domain Adaptation

Leveraging the denoising feature technique to learn from small unlabelled data helps mitigate the domain adaption challenge due to using source data and target data without labels. To compute the denoising feature for domain adaption:

Let $Y_t = (Y_1, \ldots, Y_{ns})$ represent the set of source domains data, which has a corresponding unlabelled target domain data denoted as $X^t$. Subsequently, the investigation further employs the marginalised denoising autoencoder technique to reconstruct the input data to yield the optimal reconstruction *W*, which is the weights during the reconstruction. The proposed method, therefore, minimises the loss as:

$$l(W, y) = \frac{1}{k}\sum_{k=1}^{k} \| y - \grave{y}_k w \|^2 + \omega \| w \|^2 \quad [4]$$

---

[4] https://archive.ics.uci.edu/ml/datasets.php?format=&task=&att=&area=comp&numAtt=&numIns=&type=text&sort=nameUp&view=table

[5] http://www.ai.mit.edu/projects/jmlr/papers/volume5/lewis04a/lyrl2004_rcv1v2_README.htm

[6] https://archive.ics.uci.edu/ml/datasets/Health+News+in+Twitter

Where $\grave{y}_k \in \mathbb{R}^{nxd}$ denotes the k-th denoised version of $Y = [Y^s, Y^t]$ with a random representation dropout, given the probability p, $w \in \mathbb{R}^{nxd}$ and at the same time $\omega \| w \|^2$ denotes the regularisation term or entity.

## 2.3 Experimental Results

Table 4 illustrates the dataset distribution of the Reuters-21578 datasets used for the pretraining of the model, with the number entitles, the number of linkable entities and entities that were not linked denoted as number of nil entities for training.

Table 4. The Distribution of the Dataset Statistics

| | **Number of Entities** | **Number of Linkable Entities** | **Number of Nil Entities** |
|---|---|---|---|
| Training | 3850 | 3265 | 394 |
| development | 610 | 380 | 290 |
| Test | 2908 | 1681 | 701 |

Following the chosen pretraining technique, the study compares it with the raw datasets without pretraining. Table 5 shows the Rand index average score for ten-run experiments for both pretrained and not pretrained.

Table 5. Rand index score for all the 10 run experiments

| **No.** | **Not Pretrained** | **Pretrained** |
|---|---|---|
| 1 | 68.61 | 95.15 |
| 2 | 71.22 | 96.43 |
| 3 | 66.81 | 95.63 |
| 4 | 70.30 | 95.20 |
| 5 | 70.00 | 97.35 |
| 6 | 68.82 | 97.21 |
| 7 | 72.50 | 96.55 |
| 8 | 70.01 | 96.07 |
| 9 | 60.41 | 95.55 |
| 10 | 67.71 | 96.18 |
| **Mean ± SD** | **72.50±60.41** | **97.35±95.15** |

### 2.3.1 Denoising Autoencoder Architecture

The study analyses the performance of the two convolutional denoising autoencoder architectures given the depth *d* = (1, 2, 3, 4, 5, 6, 7) to determine the best representation learning architecture Figure 3. Experiment 1 **(left)** is the variation of cluster accuracy: Accuracy measure of Convolutional Denoising Autoencoder **(A)**-top and Convolutional Denoising Autoencoder **(B)-**bottom. Experiment 1 **(right)** is the variation of cluster accuracy after an increase in d**:** Accuracy measure of Convolutional Denoising Autoencoder **(A)-top** and Convolutional Denoising Autoencoder **(B)-**bottom. Figure 3, further illustrates the cluster accuracy variation against the increase in *d* for both experiments, using 10-fold-cross validation with an average score reported. Experiment 1(**left**) denotes the variation of cluster accuracy, while experiment 2 (**right**) is the variation of cluster accuracy after an increase in *d.* The value of *d* represents the number of kernels at the encoding layer, where *d* also determines the number of representations learned for entity mentions, which significantly influences cluster accuracy. As shown in Figure 3, the cluster is comparatively lower when the value of *d* is less or equal to 3 but stabilises after 3.

The results in Figure 3 also demonstrates that an increase in the value of *d* does not essentially facilitate an increase in the cluster accuracy. The highest cluster accuracy for the first experiment (left of **Error! Reference source not found.**) shows that the convolutional denoising autoencoder $\mathbf{A}_{o=X'}^{k=7}$= 0.908 and $\mathbf{B}_{o=X'}^{k=7}$ = 0.907. The cluster accuracy's second experiment (right of Figure 3) also shows that the convolutional denoising autoencoder $\mathbf{A}_{o=X'}^{k=7}$**=** 0.897 and the convolutional autoencoder $\mathbf{B}_{o=X'}^{k=7}$= 0.914. Where *k* is the number of kernel and *o*, the output dimension and given that the experiment is unsupervised, the output dimension could not be predetermined, it is therefore denoted as $X'$. Figure 3 further shows the best entity mention activity occurs when *d =5* and *d= 4* for Convolutional denoising autoencoder **A** and **B,** respectively.

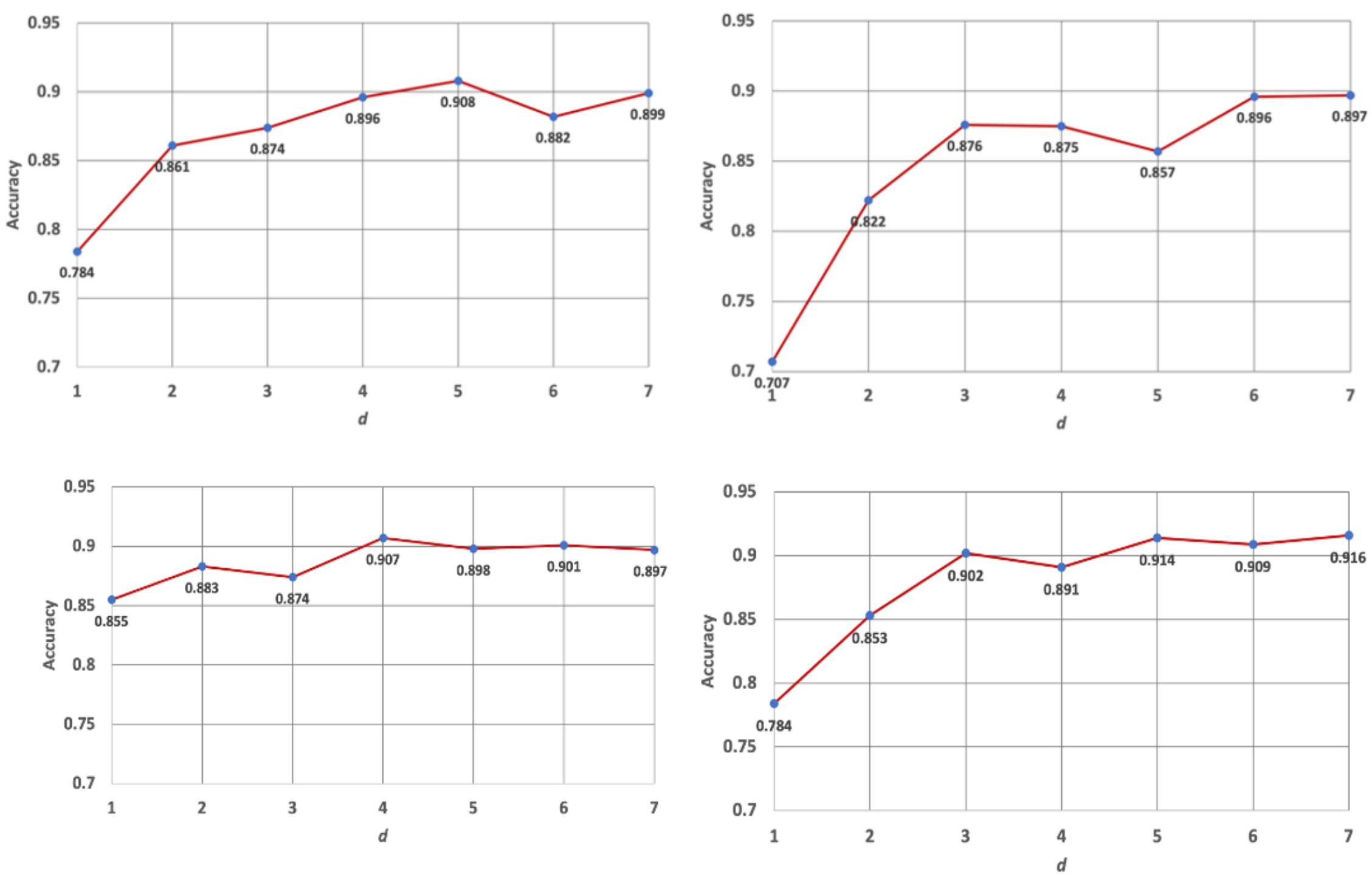

*Figure 3. Experiment 1* **(left)** is the *variation of cluster accuracy*: Accuracy measure of Convolutional Denoising Autoencoder **(A)**-top and Convolutional Denoising Autoencoder **(B)-**bottom. Experiment 1 **(right)** is *the variation of cluster accuracy after an increase in d*: Accuracy measure of Convolutional Denoising Autoencoder **(A)-top** and Convolutional Denoising Autoencoder **(B)-**bottom.

## 2.3.2 Evaluation of Clustering Performance

The experiment in this section analyses the proposed approach's cluster performance of using convolutional denoising autoencoders in closely related entity mentions. The investigation in this section further performs five-fold cross-validation to evaluate the cluster performance. Figure *4* is a boxplot of the accuracies from the five distributions of the data. The middle quartile of the first fold shows an average accuracy score of about 0.65. However, the middle quartile for the second, third and fourth fold is consistent at an average accuracy of about 0.6. The fifth middle quartile on the other hand shows an average accuracy of about 0.58. The proposed convolutional denoising autoencoder performs well in most of the cluster activities observed during the experiment.

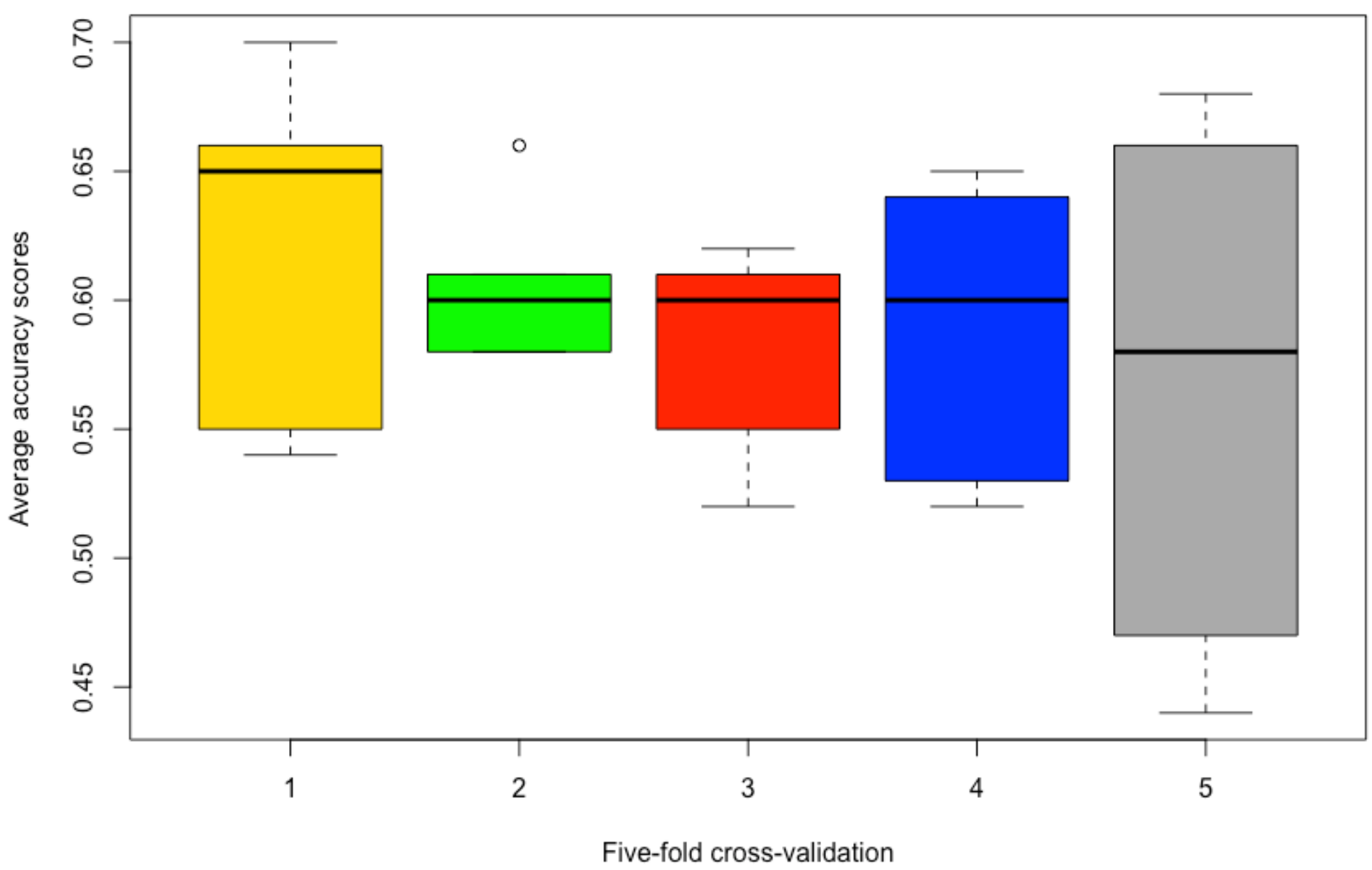

Figure 4. Five-fold cross-validation with the average accuracy scores from the five folds

A further test of significance on the average 5-fold cross validation using paired Chi-square shows that X-squared = 0.040078, df = 16, p-value = 1. Where the X-squared measures the difference between the observed and the average output scores, thus the significance level. The 0.04 score denotes the level of significance for the 5-fold cross validation, with a threshold of 0.5 indicating the highest level of significance of 5% risk establishing the correlation between the output data.

The degree of freedom (df) of 16 represents the maximum number of logically independent values and with the freedom to vary in the data sample. The p-value of 1 in this case signify that the occurrences of the 5-fold cross validation values are purely due to chance, whereas the least value of 0 would have implied the observed values are unlikely due to chance (Table *6*)

Table 6. The Average scores of the 5-fold cross validation

| Folds | First | Second | Third | Fourth | Fifth |
|---|---|---|---|---|---|
| 1 | 0.55 | 0.58 | 0.52 | 0.53 | 0.44 |
| 2 | 0.54 | 0.6 | 0.55 | 0.52 | 0.47 |
| 3 | 0.65 | 0.58 | 0.61 | 0.6 | 0.58 |
| 4 | 0.66 | 0.61 | 0.62 | 0.64 | 0.66 |
| 5 | 0.7 | 0.66 | 0.6 | 0.65 | 0.68 |

Table 7. Comparison of proposed methods on cluster accuracy

| Architecture | Cluster accuracy | SD of count per cluster |
|---|---|---|
| Convolutional Denoising Autoencoder (Accelerometer) A | 0.82 | 8.73 |
| Convolutional Denoising Autoencoder (Accelerometer) B | 0.85 | 8.78 |

Table 7 shows a comparison of the average cluster accuracy and the standard deviation (SD) of count per cluster of the proposed techniques. The study further analyses the effect of the kernel size on the performance of the entity mentioned activity. When a small kernel is introduced to the proposed method, it allows it to learn quality complex representations from the data, while when a large kernel is used, more general representations are learned. When the convolutional layers and the kernel size of **B** are varied, the convolutional denoising autoencoder captured more information or patterns in the data than when **A** is varied**.** On the other hand,

Table 9 illustrate the cluster average accuracy of the convolutional denoising autoencoder for **B** with varied kernel sizes. The kernel size did not significantly impact the entity mentioned activities. Similarly, smaller kernel sizes produce a slight decrease in cluster accuracy. Nevertheless, a small kernel size helps reduce the number of weights and eventually the computational cost.

Table 8 Comparing cluster accuracy of convolutional autoencoder using accelerometer with different kernel sizes(A)

| Kernel Size | Average Cluster accuracy |
|---|---|
| 3, 5, 7, 9, 11 | 0.935 |
| 3, 3, 5, 7, 9 | 0.928 |
| 3, 3, 3, 5, 7 | 0.918 |

Table 9.Comparing cluster accuracy of convolutional autoencoder A using accelerometer with different kernel sizes

| Kernel Size | Average Cluster accuracy |
|---|---|
| 3, 5, 7, 9, 11 | 0.931 |
| 3, 3, 5, 7, 9 | 0.926 |
| 3, 3, 3, 5, 7 | 0.912 |

### 2.3.3 Unsupervised Representation Transfer and Domain Adaptation

In this section, the experiment focuses on the transfer between the source and the target domain. Most transfer learning models require a specific domain with a large, labelled dataset to transfer knowledge and learn representation from a new dataset (Chang et al., 2017; Michau & Fink, 2021). However, previous experiments in (Fianyi, 2024) show that unsupervised pretraining helps knowledge transfer well on different assumptions and across domains with unlabelled datasets. While different techniques (few-short-learning, domain adaptation, zero-shot learning, and others) (Artetxe & Schwenk, 2019; Li et al., 2020; Michau & Fink, 2021) have been used successfully for transfer learning, to the best of the researchers' knowledge in this study, there are no reported works that study unsupervised transfer learning where a different unlabelled small dataset is introduced to an already trained model.

Therefore, the experiment in this section proposes a transfer cluster mechanism that learns *discriminative* hidden space on small unlabelled data, a knowledge transfer from a pretrained word embedding model, and a knowledge-based benchmark dataset for the related task. The investigation leverage pairwise to transfer word embedding to extract entity mentions from the dataset, which requires that words and knowledge base resource are represented in the same space to support a more *discriminative* clustering; the ACE2005 dataset has these characteristics. To establish the condition for a small dataset, the study considers a number of subsets from the original datasets, 20%, 30%, and 40%, thus less than half of the original dataset. The selected subsets are based on the quantity of the datasets that adapts well with the proposed model after initial investigation.

To avoid accidental bias in the selection of the subset of the data, the study adopts an unsupervised *aggregation* technique (Ulan et al., 2021) that is based on the intrinsic properties of the unlabelled training data, thus the cumulative probability distribution of the independent variables and their related dependencies. The aggregation model helps determine how well the output represents properties of the input tuples. This approach has been used in Zhuang et al. (2021) performed unsupervised representation learning using a centralised data from the internet using an aggregated encoder to segregate the datasets to learn the visual representation of the data distribution. Kersten et al. (2021) used the aggregated techniques for a semantic crisis related tweets, where they randomly divided tweets from the past and current tweets under supervised learning conditions.

The aggregation maps many input variables to a single input variable. The study assumes that the size of the input variable is fixed say *q.* The study also assumes that all input and output variables have unit interval [0, 1]. Furthermore, the function that maps the *q*-dimensional unit cube to the unit interval:

$$R: [0,1]^q \longmapsto [\,0,1] \qquad [5]$$

This becomes an aggregation when the monotonicity and the boundary conditions properties are satisfied.

Monotonicity is therefore denoted as:

$$R(a_1, \dots, a_q) \leq R(b_1, \dots, b_q) \Leftrightarrow \bigwedge_{i \in \cdots q} a_i \leq b_i \qquad [6]$$

And the boundary condition is represented as:

$$R(0, \dots, 0) = 0 \wedge R\,(1, \dots 1) = 1 \qquad [7]$$

Then the study applies the principal component analysis proposed in(Fianyi et al., 2024), to perform dimensionality reduction of the embedding elements while training the aggregated data from a high-dimensional vector space (this is because the aggregated data showed features of high-dimensionality during the initial investigation) to a low-dimensional vector space.

The investigation evaluates the effectiveness of clustering entity mentions from small unlabelled data with the help of pretraining on a knowledge base resource while comparing it with other existing two benchmark cluster models. Firstly, the proposed methods showed the possibility of learning more transferrable representations than the conventional transfer learning approaches. Table 10 illustrates the comparison of the ACE2005 datasets used for transfer learning with the state-of-the-art transfer learning datasets.

The ACE2005, Health News in Twitter and the TERRY datasets were all pretrained under the same conditions with the same parameters. They were all segregated into the number of tokenised development or training set, the number of transferable entities based on ground truth set and the number of validations set used to verify and validate the proposed model. The quantity of the ACE2005 set (from the development, transfer, and validation number set) used for training the proposed transfer learning model is lesser than the benchmark sets. The significance of this is to ascertain as well as evaluate how the proposed pretraining models perform when other reduced/small datasets are introduced, thus in a range of 20%, 30% and 40%. Contextually, all the three datasets belong to different domains, nevertheless, the underpinning commonality is the fact that they are unlabelled textual corpus and fit for transfer learning in a downstream named entity linking task.

Table 10. Comparative evaluation of the ACE2005 dataset with transfer learning benchmark datasets

| | | **(20%)** | | | **(30%)** | | | **(40%)** | | |
|---|---|---|---|---|---|---|---|---|---|---|
| **Datasets** | **Domain** | **Dev. No** | **Transfer** | **Validation No** | **Dev. No** | **Transfer** | **Validation No.** | **Dev. No.** | **Transfer** | **Validation No.** |
| TERRY | Text preprocessing | 43406 | 18000 | 2766 | 47747 | 19800 | 3043 | 52522 | 21780 | 3347 |
| Health News in Twitter | Health text | 11600 | 8000 | 1980 | 12760 | 8800 | 2178 | 14036 | 9680 | 2396 |
| **ACE 2005** | Linguistics | 2808 | 1608 | 601 | 3089 | 1769 | 661 | 3398 | 1769 | 727 |

### 2.3.4 The Impact of joint application of unsupervised pretraining and unsupervised transfer learning algorithms

Throughout the experiment, one of the common problems with the transfer learning scenario is the distribution discrepancies between source domain data and target domain data, which is characterised by differences in distribution and dimensionality of features. Several studies (Long et al., 2017; Noori Saray & Tahmoresnezhad, 2022; Wang & Carbonell, 2018; Zhou, 2022) have also shown the challenges of distributional discrepancies in transfer learning, making it consistent with the investigation of this study. The proposed method evaluates the effect of unsupervised pretraining on the performance of unsupervised transfer learning algorithms to mitigate the distribution discrepancy problem and enhances feature learning for entity linking.

This investigation focuses on:

1. Obtaining projections for both the source and target domain data to help account for different conditional distributions and a reduction in the domain discrepancy.
2. Minimising the cluster error on a new representation of source domain data while finding best representation for clustering.
3. Increasing the manifold consistency that underpins the marginal distributions for the source and target domain data.

The proposed method introduces a joint statistical alignment to modify a joint subspace model for transfer learning. The proposed method adopts conditional distributions[7] for marginal projection of both source and target domain data. Subsequently, the study defines Equation 5.8 to obtain a new representation after pretraining the source domain data (s) for the target domain data (t).

$$\min_{S\,T} qr \left( [\, S^q T^q \begin{bmatrix} M_s & M_{st} \\ M_{ts} & M_t \end{bmatrix} \begin{pmatrix} S \\ T \end{pmatrix} \right) \quad [8]$$

Where

$$M_s = Y_s((1-\gamma)P_s + \gamma \sum_{c=1}^{c} P_s^c)Y_S^T \quad [9]$$

To minimise cluster error to learn new representation during unsupervised pretraining for the unsupervised transfer learning task, the proposed methods map the original data through S and T to discover the new representation of both source and target domain data. The goal is to learn a classifier ***f*** that adapts during the unsupervised pretraining stage on the unlabelled source domain

[7] Conditional distribution encompasses the distribution of values for a respective variable when the value of the other variable is being specified, thus the variables for the source and target domain.

$S_d$ *for* the classification of the target domain. Additionally, to minimise the structural risk function and learn ***f*** during pretraining, the study proposes:

$$f = arg \min_{f \epsilon k} \sum_{j=1}^{n} \ell(f(X_{si}), y_j + \eta \parallel f \parallel_k^2 \quad [10]$$

Where $f \epsilon k$ constitutes the classifier that generates the kernel Hilbert space and $\eta$ denotes the parameter for the regulariser, $\ell$ represents the loss function squared $\ell = (y_{j-} f(X_{si}))^2$ that is used to measure the performance of the classifier *f* on predicting the entities from the unlabelled data. The idea behind the proposed transfer learning approach is to automatically infer from the dimensionality of the vector space to align as much as possible the subspace of the source and target domain.

Consequently, the study computes the principal component analysis of both datasets ($PCA_S$ and $PCA_T$) in the vector space. Subsequently, the study combines the datasets into one to enable the computation of the sub vector space $PCA_{S+T}$. The metrics to capture this notion of combining the dataset in one vector is defined regarding the principal angle in the subspace classification of the Grassmannian is defined as:

$$D(d) = 0.5[\sin \alpha_d + \sin \beta_d ] \quad [11]$$

Where $\alpha_d$ represents the $d^{th}$ principal angle between $PCA_S$ and $PCA_{S+T}$ then the $\beta_d$ is between the $PCA_T$ and $PCA_{S+T}$. Notably, either $\sin\alpha_d$ or $\sin\beta_d$ is known as the minimum correlation distance (Huang et al., 2018). The D(d) is at maximum 1. A small value implies that both $\alpha_d$ and $\beta_d$ is also small, also indicating that $PCA_S$ and $PCA_T$ are ranged at the $d^{th}$ dimension. When the optimal value is 1, the subspaces for $PCA_S$ and $PCA_T$ have orthogonal directions, thus $\alpha_d = \beta_d = \pi/2$. At this point, domain adaptation becomes difficult because the variances captured in a single sub vector space would not transfer to the other sub vector space. To identify the best value of *d,* the study employs the greedy approach where:

$$d^* = \min\{d|D(d) = 1\} \quad [12]$$

Naturally, the best value of d* is expected to be at the highest point, this helps to maintain the variances in the source domain to build a good cluster but not too high that the two sub spaces do not have be in in orthogonal direction.

To increase the manifold consistency in a typical transfer learning that underpins the marginal distributions for the source and target domain data. The approach maximises the regularity in the inherent manifold structure of the data to predict the structure of the source and domain data using:

$$M_f(R_s, R_t) = \sum_{i,j=1}^{n+p} Q_{ij} \,(f(X_i) - f(X_j))^2 \quad [13]$$

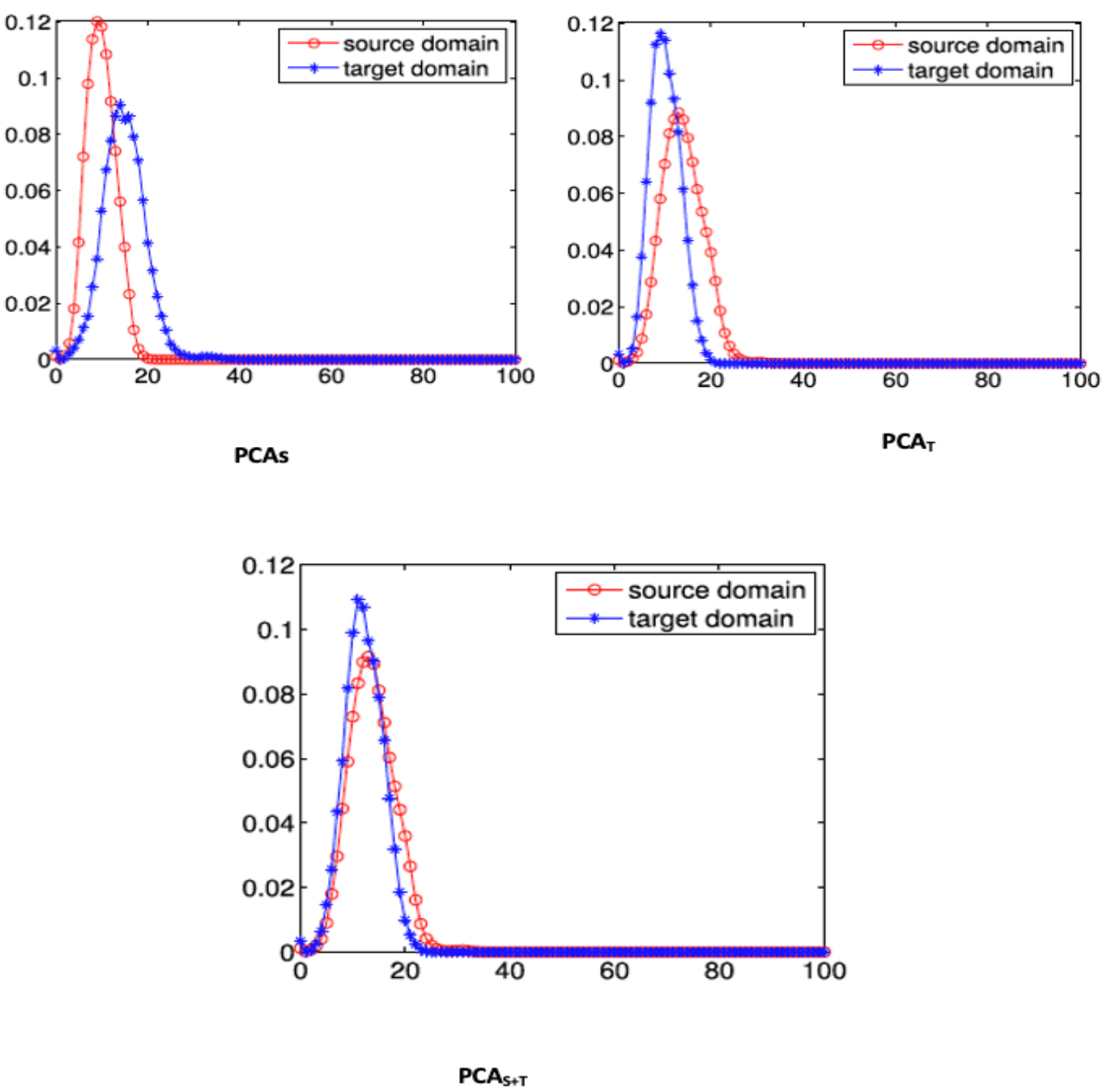

Figure 5. A discriminative pairwise to show the distance between each domain (source and target domain) with regards to four different subspaces. The proposed method induces a subspace that makes the difference between the histogram of the source domain, and the target domain is negligible

Figure 5 shows the impact of unsupervised pretraining techniques for domain adaptation using the pairwise discriminative techniques with more than one sub vector spaces. The source domain data is Reuters21578, and ACE2005 is the target domain data. The subspace shows that the source and target domain data are drawn closer, and Table 11 statistically proves this assertion. This empirical evidence denotes that unsupervised pretraining help minimises distortion and cluster error and promotes consistency in the source domain data for the target domain data in unsupervised transfer learning. Subsequently, it supports the performance of the discriminative transfer learning approach preceded by unsupervised pretraining on the source domain. The experiment has also proven that unsupervised pretraining improves strong link match in an unsupervised transfer learning task with small unlabelled datasets.

Table 11. The ratio (in percentage) of the distance using three subspaces within the source and the target domain data

| Domain | $PCA_S$ | $PCA_T$ | $PCA_{S+T}$ |
|---|---|---|---|
| Reuters-21578 | 9.76 | 8.68 | 6.49 |
| ACE2005 | 19.81 | 16.80 | 15.8 |
| Average | 14.79 | 12.74 | 11.15 |

### 2.3.5 Performance Comparison with the start-of the-art Approach

The work in this section compares the proposed transfer learning techniques for NEL with the current state-of-the-art solutions. The proposed unsupervised transfer learning technique preceded by unsupervised pretraining is promising in several respects. Table 12 shows the report of the consistency of the proposed methods with the different datasets from different domain. The table shows that the proposed Unsupervised Transfer Learning (**UTL)** approach has consistent performance with other benchmark datasets. The study compares the performance of the model based on various subsets of the original datasets 20%, 30% and 40%. The study also compares the model basis for transfer learning against the proposed combination of discriminative and k-means model to learn representation from small datasets.

Table 13 shows the evaluation of the proposed method (discriminative + k-means) with the ACE2005 datasets compared with benchmark transfer learning datasets. The experiment used three different subsets. While k-means clustering techniques and DCANS have been used over the years successfully in various transfer learning tasks, employing these algorithms for small datasets is still new. The results in the Table 13 shows that a combination of discriminative algorithms with k-means have shown promising signs in unsupervised transfer learning. For the ACE2005 dataset with the discriminative + k-means and at the data subset of 20%, the F1-measure saw 67.5, and at 30% the F1-measure saw 71.4 and 78.7 when the subset data was 40%. For all the three subset data categories, the ACE2005 dataset with the proposed discriminative + k-means techniques saw the highest percentage of values indicating a better performance against the benchmark datasets and their associated model basis in unsupervised transfer learning tasks.

Table 13. Comparison of the ACE2005 datasets with benchmark datasets measured against validation accuracy of transfer learning on three different subsets of the original datasets

| | | 20% | | | 30% | | | 40% | | |
|---|---|---|---|---|---|---|---|---|---|---|
| Datasets | **Model Basis** | **Precision** | **Recall** | **F1 Measure** | **Precision** | **Recall** | **F1 Measure** | **Precision** | **Recall** | **F1 Measure** |
| TERRY | DBSCAN | 65.2 | 59.4 | 62.2 | 68.7 | 63.3 | 65.9 | 73.8 | 69.2 | 71.9 |
| Health News in Twitter | K-means | 70.3 | 63.5 | 66.7 | 75.1 | 65.4 | 70.0 | 80.6 | 72.0 | 76.1 |
| **ACE 2005** | **Discriminative +k-means** | 75.3 | 61.1 | **67.5** | 78.0 | 65.8 | **71.4** | 84.5 | 73.7 | **78.7** |

Notably, the benchmark datasets rely on knowledge bases to transfer knowledge onto the new datasets. However, the proposed model for this investigation relies on unsupervised pretraining techniques to transfer knowledge to the new datasets. In the CoDAE proposed for the activity of the entity mentioned, the CoDAE consist of fully connected three hidden layers with 300 units in the first hidden layer, 200 in the second hidden layer and then 50 units in the last hidden layer. Nevertheless, the current transfer learning techniques for named entity linking and their related tasks used a minimum of 1000 units with only two layers as inspired by Li et al. (2019) and . This suggests that more layers help representation learning, especially with small datasets.

## 2.4 Summary

This study proposes a combination of unsupervised pretraining and unsupervised learning transfer learning techniques for unsupervised Named Entity Linking with small dataset. The study proposes Convolutional Denoising Autoencoders (CoDAE) to automatically extract knowledge representation with discriminative/k-means technique under an accelerometer so that there will be no need for manual feature engineering. The proposed method also adapts convolutional pooling layers to leverage the local structure of the small dataset. Additionally, the study proposes unsupervised pretraining to learn salient word embedding from a small unlabelled dataset. Furthermore, the study investigates how unsupervised pretraining can help transfer knowledge or embedding from a source domain data to a target domain data in a transfer learning scenario without external knowledge bases. The investigations reveal that unsupervised pretraining helps augment small datasets, and when used to learn word embedding, it improves semantic similarity and captures new named entities in a different dataset from the ones used in training the network.

One of the main challenges with unsupervised transfer learning is that every word and its meaning must be encoded in a single space to learn quality word embedding representation for the respective task. This challenge causes some words to be placed in the wrong position triggering a false positive entity, where entities are in the proposed target link but not in the ground truth domain. The proposed model with three subset categories compared with other benchmark datasets shows a cluster accuracy of 67.5 with a 20% for ACE2005 as against 66.7 for health news in twitter and 62.2 for TERRY. Similarly, when the sub dataset is further increased to 30%, The transfer cluster accuracy for ACE2005 shows 71.4, 70.0 for health news in twitter and 65.9 for TERRY. It is not surprising the model performed better when the sub dataset was increased to 40%, where the F-measure with ACE2005 saw 78.7, 76.1 with health in news data and 71.9 with TERRY. While more data shows better performance, the scenario of learning from small dataset is still critical and possible as demonstrated in this study. The study did not consider live feed data, which is necessary in leveraging real-world datasets. Future studies could consider live feed unlabelled datasets and explore the representation of each word occurrence with its respective meaning during pretraining to facilitate real-time transfer learning.

Arnold, A., Nallapati, R., & Cohen, W. W. (2007). A comparative study of methods for transductive transfer learning. Seventh IEEE international conference on data mining workshops (ICDMW 2007),

Artetxe, M., & Schwenk, H. (2019). Massively multilingual sentence embeddings for zero-shot cross-lingual transfer and beyond. *Transactions of the Association for Computational Linguistics*, *7*, 597–610.

Ashraf, S., Brabyn, L., & Hicks, B. J. (2013). Introducing contrast and luminance normalisation to improve the quality of subtractive resolution merge technique. *International Journal of Image and Data Fusion*, *4*(3), 230–251.

Boudiaf, M., Kervadec, H., Masud, Z. I., Piantanida, P., Ben Ayed, I., & Dolz, J. (2021). Few-shot segmentation without meta-learning: A good transductive inference is all you need? Proceedings of the IEEE/CVF Conference on Computer Vision and Pattern Recognition,

Cao, Z., You, K., Long, M., Wang, J., & Yang, Q. (2019). Learning to transfer examples for partial domain adaptation. Proceedings of the IEEE/CVF conference on computer vision and pattern recognition,

Chadha, G. S., Islam, I., Schwung, A., & Ding, S. X. (2021). Deep Convolutional Clustering-Based Time Series Anomaly Detection. *Sensors (Switzerland)*, *21*(16), 5488.

Chang, H., Han, J., Zhong, C., Snijders, A. M., & Mao, J.-H. (2017). Unsupervised transfer learning via multi-scale convolutional sparse coding for biomedical applications. *IEEE Transactions on Pattern Analysis and Machine Intelligence*, *40*(5), 1182–1194.

Dai, X., Bikdash, M., & Meyer, B. (2017). From social media to public health surveillance: Word embedding based clustering method for twitter classification. SoutheastCon 2017,

Duong, T. N., Doan, N. N., Do, T. G., Tran, M. H., Nguyen, D. M., & Dang, Q. H. (2022). Utilizing Half Convolutional Autoencoder to Generate User and Item Vectors for Initialization in Matrix Factorization. *Future Internet*, *14*(1), 20.

Fianyi, I. (2024). *Unsupervised deep learning approach for information extraction* University of Tasmania].

Fianyi, I., Montgomery, J., & Yeom, S. (2024). Unsupervised Pretraining Approach for Open Relation Extraction.

Hemavathi, D., Srimathi, H., & Sornalakshmi, K. (2019). A Hybrid Technique for Unsupervised Dimensionality Reduction by Utilizing Enriched Kernel Based PCA and DBSCAN Clustering Algorithm. International Conference on Inventive Computation Technologies,

Huang, W.-b., & Sun, F.-c. (2016). Building feature space of extreme learning machine with sparse denoising stacked-autoencoder. *Neurocomputing*, *174*, 60–71.

Huang, Z., Wu, J., & Van Gool, L. (2018). Building deep networks on grassmann manifolds. Proceedings of the AAAI Conference on Artificial Intelligence,

Kersten, J., Bongard, J., & Klan, F. (2021). Combining Supervised and Unsupervised Learning to Detect and Semantically Aggregate Crisis-Related Twitter Content. *Analysis of Detection Models for Disaster-Related Tweets*, 744–754.

Li, J., Ye, D., & Shang, S. (2019). Adversarial Transfer for Named Entity Boundary Detection with Pointer Networks. IJCAI,

Li, Z., Yao, H., & Ma, F. (2020). Learning with small data. Proceedings of the 13th International Conference on Web Search and Data Mining,

Long, M., Zhu, H., Wang, J., & Jordan, M. I. (2017). Deep transfer learning with joint adaptation networks. International conference on machine learning,

Mengliev, D., Barakhnin, V., Abdurakhmonova, N., & Eshkulov, M. (2024). Developing named entity recognition algorithms for Uzbek: Dataset Insights and Implementation. *Data in Brief*, *54*, 110413.

Mesnil, G., Dauphin, Y., Glorot, X., Rifai, S., Bengio, Y., Goodfellow, I., Lavoie, E., Muller, X., Desjardins, G., & Warde-Farley, D. (2012). Unsupervised and transfer learning challenge: a deep learning approach. Proceedings of ICML Workshop on Unsupervised and Transfer Learning,

Michau, G., & Fink, O. (2021). Unsupervised transfer learning for anomaly detection: Application to complementary operating condition transfer. *Knowledge-Based Systems*, *216*, 106816.

Niepert, M., Ahmed, M., & Kutzkov, K. (2016). Learning convolutional neural networks for graphs. International conference on machine learning,

Noori Saray, S., & Tahmoresnezhad, J. (2022). Iterative joint classifier and domain adaptation for visual transfer learning. *International Journal of Machine Learning and Cybernetics*, *13*(4), 947–961.

Qu, X., Gu, Y., Xia, Q., Li, Z., Wang, Z., & Huai, B. (2023). A survey on arabic named entity recognition: Past, recent advances, and future trends. *IEEE Transactions on Knowledge and Data Engineering*.

Qu, X., Zeng, J., Liu, D., Wang, Z., Huai, B., & Zhou, P. (2023). Distantly-supervised named entity recognition with adaptive teacher learning and fine-grained student ensemble. Proceedings of the AAAI Conference on Artificial Intelligence,

Raffel, C., Shazeer, N., Roberts, A., Lee, K., Narang, S., Matena, M., Zhou, Y., Li, W., & Liu, P. J. (2020). Exploring the limits of transfer learning with a unified text-to-text transformer. *J. Mach. Learn. Res.*, *21*(140), 1–67.

Shi, P., Ye, W., & Qin, Z. (2021). Self-Supervised Pre-training for Time Series Classification. 2021 International Joint Conference on Neural Networks (IJCNN),

Tekumalla, R., & Banda, J. M. (2021). Using weak supervision to generate training datasets from social media data: a proof of concept to identify drug mentions. *Neural Computing and Applications*, 1–9.

Ulan, M., Löwe, W., Ericsson, M., & Wingkvist, A. (2021). Aggregation as Unsupervised Learning and its Evaluation. *arXiv preprint arXiv:2110.15136*.

Wan, Z. Y., Dodov, B., Lessig, C., Dijkstra, H., & Sapsis, T. P. (2021). A data-driven framework for the stochastic reconstruction of small-scale features with application to climate data sets. *Journal of Computational Physics*, 110484.

Wang, Z., & Carbonell, J. (2018). Towards more reliable transfer learning. Joint European Conference on Machine Learning and Knowledge Discovery in Databases,

Zhou, C. (2022). *Mitigating Semantic and Distributional Discrepancies in Natural Language Processing* University of Washington].

Zhu, Y., Wu, X., Qiang, J., Hu, X., Zhang, Y., & Li, P. (2022). Representation learning with deep sparse auto-encoder for multi-task learning. *Pattern Recognition*, *129*, 108742.

Zhuang, W., Gan, X., Wen, Y., Zhang, S., & Yi, S. (2021). Collaborative unsupervised visual representation learning from decentralized data. Proceedings of the IEEE/CVF International Conference on Computer Vision,